\documentclass[10pt]{article}
\usepackage{rev_sbrn}
\usepackage{times,amsfonts,enumerate,amssymb,amsmath,epsfig,subfigure,bm,cite}
\usepackage[applemac]{inputenc}
\usepackage[OT1]{fontenc}
\usepackage[brazil]{babel}
\usepackage{algorithm2e}

\usepackage[a4paper, 
hmargin={2cm,1cm}, 
vmargin={2cm,2cm},
footskip=5mm]{geometry}

\begin{document}
\thispagestyle{empty}
\title{AUTOMATIC SYSTEM FOR COUNTING CELLS WITH ELLIPTICAL SHAPE}

\author{
{\bf Wesley Nunes Gon\c{c}alves, Odemir Martinez Bruno}\\
{\normalsize Instituto de F\'{i}sica de S\~{a}o Carlos - Universidade de S\~{a}o Paulo} \\ 
{\normalsize Av. Trabalhador S\~{a}o-carlense, 400 - CEP: 13560-970} \\ 
{\normalsize Cx. Postal 369 - S\~{a}o Carlos - SP - Brasil} \\ 
{\normalsize wnunes@ursa.ifsc.usp.br, bruno@ifsc.usp.br}  \\ \\
%
%
}
\maketitle 
\thispagestyle{empty}


\noindent{}{\bf\large Abstract --} 
This paper presents a new method for automatic quantification of ellipse-like cells in images, an important and challenging problem that has been studied by the computer vision community.
The proposed method can be described by two main steps.
Initially, image segmentation based on the k-means algorithm is performed to separate different types of cells from the background.
Then, a robust and efficient strategy is performed on the blob contour for touching cells splitting.
Due to the contour processing, the method achieves excellent results of detection compared to manual detection performed by specialists.
\\

\noindent{}{\bf\large Keywords --}
Touching cells splitting, computer vision, pattern recognition.
\\

\renewcommand{\figurename}{Figure}
\renewcommand{\tablename}{Table}
\Section{Introduction}
\label{sec:intro}
Image analysis methods for identifying and quantifying objects (e.g. blood cell, bacteria, nanostructure) are an essential task for many research areas.
In microbiology, for instance, examining and quantifying cells by microscopy has been a central method for studying cellular function, such as the estimation of parasitemia from microscopy images of blood \cite{Halim2006} and the quantification of cell adhesion for understanding physiological phenomena.
Quantifying cells in images becomes even more important because in most cases, the sequence of the research depends on the results obtained in this step.

Usually, cell counting is performed in a manual process, which takes hours or even days of work.
Due to human factors such as fatigue and distraction, the results obtained by manual counting are not completely reliable or reproducible.
Thus, automation of this process has attracted increasing attention from computer vision community.
Besides providing more reliable and reproducible results, automatic cell counting also provides statistics of the cells that a human being is unable to estimate, as area, perimeter and volume.

Several methods for counting cells in images have been proposed in the literature.
A large portion of the methods is based on the watershed algorithm~\cite{Mao2006,Peter2008,Vincent1991}, whose basic idea is to flood an image as a topography relief.
Although less explored, active contours~\cite{Eom2006,Bamford1998} and region growing methods~\cite{Ning2010,Anoraganingrum1999} have been used in several other methods and obtained interesting  results.
There are methods that counting cells in images based on morphological operations~\cite{Ruberto2002,Rackey1999} and methods that use priori information of the object shape~\cite{Wang1998,Kumar2006}.
Although cell counting have been heavily studied by computer vision community, most methods does not provide satisfactory results for images with complex touching cells \cite{Bai2009}.

This paper proposes an approach for automatic counting of cells in images that combines k-means segmentation and ellipse fitting.
Different types of cells in the same image (e.g. cells of different colors) are segmented from the background using k-means algorithm.
After segmentation, ellipse fitting is performed on the contour of blobs to separate touching cells.
Two set of experiments were carried out using three types of cell images.
The first experiment aims to evaluate the proposed method by using images marked by specialists.
This experiment was performed in images with high density of the Lactobacillus paracasei bacteria.
These bacteria are found in human being mouth and are responsible for the majority of diseases such as caries.
Furthermore, the second experiment was performed in images containing a large number of three types of touching cells.
In both applications, the proposed approach provided excellent results compared to the manual annotation performed by a specialist.

The paper is described in four sections.
In Section 2, the proposed approach for counting cells is described from the pre-processing and segmentation of images to the ellipse fitting that provides the separation of touching cells.
Experiments and results for three types of images (images of bacteria in two stages and blood cells) are presented in Section 3.
Finally, in Section 4, conclusion and future works are discussed.

\begin{figure}[!htbp]
  \begin{center}
    \includegraphics[width=.65\columnwidth]{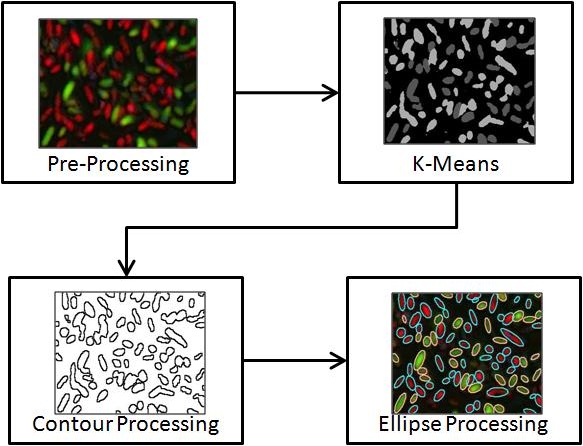}
    \caption{Summarization of the proposed approach. First, pre-processing is applied in the image. Second, cells are segmented using k-means with $k = n+1$. Then, contour of blobs are divided at concave points. Finally, an ellipse is fitted for each segment of the contour.}
    \label{fig:metodo}
  \end{center}
\end{figure}

\Section{Proposed Approach}
Proposed approach is summarized in Figure \ref{fig:metodo}.
Initially, a pre-processing is applied in order to enhance the image.
Then, cells are segmented from the background by using k-means algorithm with $k = n+1$ groups ($n$ types of cells and background).
Finally, blobs containing more than one cell are divided into segments.
These segments are determined by concave points on the contour and then an ellipse is fitted for each segment.

\subsection{Pre-processing}
In order to enhance image contrast, images are preprocessed using the decorrelation stretch method \cite{Karvelis2008}.
Decorrelation stretching method is based on principal components transformation to eliminate the correlation between bands (e.g. RGB color space).
This process involves three main steps.
First, principal component analysis is applied on the rows and columns of the image.
Then, contrast equalization is applied by a Gaussian filter.
Finally, coordination conversion is applied to the original bands.
More information can be found in \cite{Karvelis2008}.

\subsection{Segmentation using K-Means}
\begin{figure}[!htbp]
     \centering
     \subfigure[Original image.]{
           \includegraphics[width=.45\columnwidth]{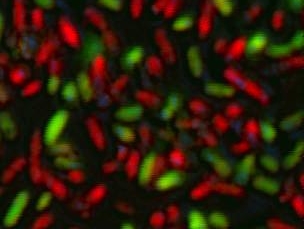}}
     \subfigure[Segmented image.]{
          \includegraphics[width=.45\columnwidth]{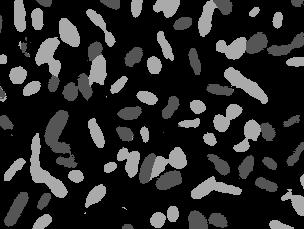}}
     \caption{Results of segmentation of cell image using k-means algorithm.}
     \label{fig:kmeans}
\end{figure}

After the pre-processing step, cells in the image are segmented from the background.
Since the images contains relevant color information (Figure \ref{fig:kmeans}(a)), segmentation is done by using the well-known k-means algorithm.
This algorithm is a clustering method that aims to partition the data into $k$ groups such that the distance between elements of the same group are minimized.
In image segmentation, each pixel is considered an element $x_i = [R_i, G_i, B_i]$ that must be assigned to one of the $k$ groups.
The algorithm has two iterative steps.
Given $k$ centroids, each pixel $x_i$ is assigned to the nearest centroid.
Then, centroids are recalculated according to the pixels belonging to each group.
The two steps above are repeated until the difference between centroids of two iterations is less than a threshold.
Figure \ref{fig:kmeans} shows an example of image segmentation using k-means algorithm with $k = 3$.

\begin{figure}[!htbp]
     \centering
     \subfigure[Original image.]{
           \includegraphics[width=.45\columnwidth]{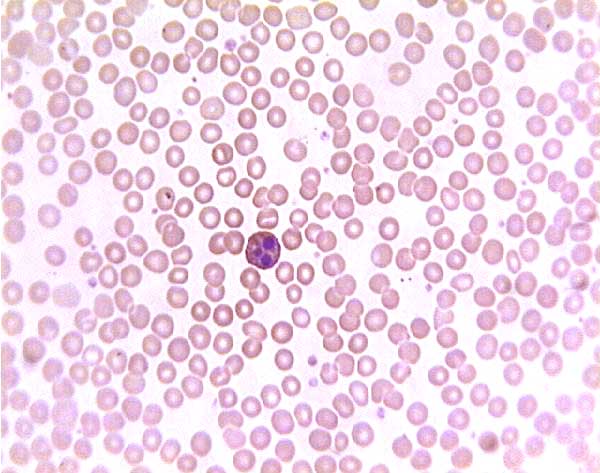}}
     \subfigure[Segmented image using k-means with $k=2$.]{
          \includegraphics[width=.45\columnwidth]{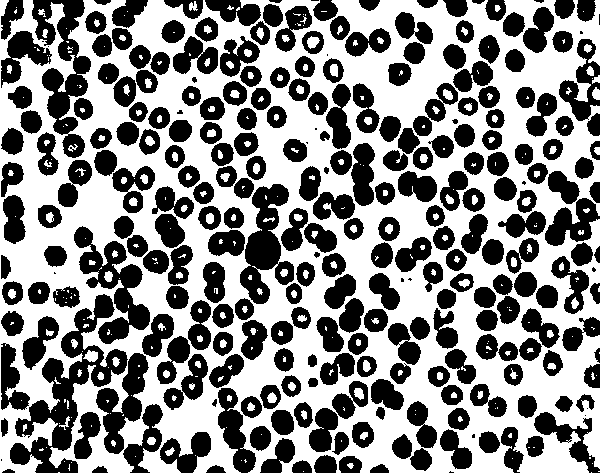}}
     \subfigure[Segmented image followed by fill hole method.]{
          \includegraphics[width=.45\columnwidth]{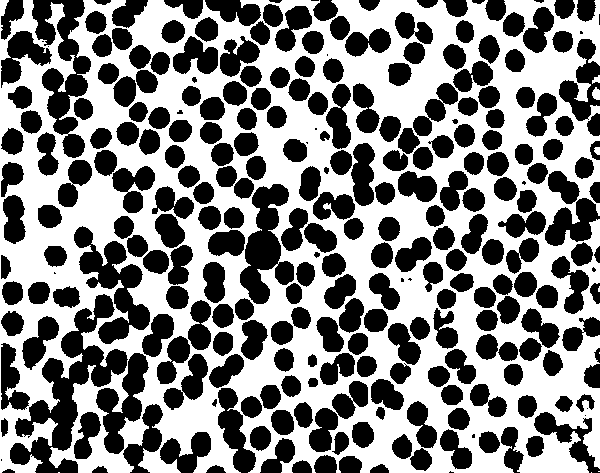}}
     \caption{Results of segmentation of blood cell image using k-means followed by the fill hole method.}
     \label{fig:nano}
\end{figure}

In some cases, the segmented image has blobs with a hole inside due to noise, image capturing or type of cell (Figure \ref{fig:nano}(b)).
To solve this problem, after the k-means segmentation, we apply a fill hole method \cite{Gonzales2006}.
Note that, if the algorithm for contour extraction is invariant to holes, this step is unnecessary.
Figure \ref{fig:nano} shows an example of the step described above.

\subsection{Contour Processing}
\begin{figure}[!htbp]
     \centering
     \subfigure[Original image.]{
          \label{fig:segmentationContourA}
          \includegraphics[width=.24\columnwidth]{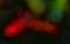}}
     \subfigure[Concave points.]{
          \label{fig:segmentationContourB}
          \includegraphics[width=.24\columnwidth]{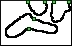}}
     \subfigure[Contour segments.]{
          \label{fig:segmentationContourC}
          \includegraphics[width=.24\columnwidth]{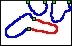}}
     \subfigure[Fitted eliipses to each segment.]{
          \label{fig:segmentationContourD}
          \includegraphics[width=.24\columnwidth]{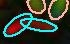}}
     \caption{Example of contour segments and ellipses.}
     \label{fig:segmentationContour}
\end{figure}

After the image segmentation, some blobs contain two or more touching cells.
In this work, the contour of the blobs is used to split touching cells.
The contour is represented by a set of points $C=\{p_{1}, p_{2}, ..., p_{n}\}$, where $p_{i} = (x,y)$ is the contour point and $n$ is the number of points.
Figure \ref{fig:segmentationContour} illustrates the contour processing problem and the separation using the proposed approach.
The main idea of the contour processing is to split the contour into segments belonging to different cells through concave points.

The original contour of the cells has many small-scale fluctuations and noises that can affect its analysis.
To decrease the influence of noise and fluctuation, a polygon approximation \cite{Wang1998} is applied to smooth the original contour $C$.
The polygon approximation provides a set of points $PAC = \{p_{1}, p_{2}, ..., p_{m}\} \mid p_{i} \in C$.
The approximation method used in this work starts with two points $p_{i}$ and $p_{j} \in C$, where $j=i+nStep$ and $nStep > 1$.
Then, distances between the line $\overline{p_{i}p_{j}}$ and each point $p_t | i < t < j$ are calculated and compared to a threshold $dTh$.
If the distance of a point $p_{t}$ is greater than $dTh$, this point belongs to the polygon approximation ($p_{t} \in PAC$), $p_{i}$ moves to $p_{t}$ and the procedure is repeated.
Otherwise, $p_{j}$ moves to the next point and the distances are recalculated until there is a point $p_{t}$ or $p_{j}$ reaches the end of the contour.
When $p_{i}$ cover all contour points, the procedure is terminated.

\begin{figure}[!htbp]
     \label{fig:concavePoints}
     \centering
     \subfigure[Illustration of concave point. An point is discarted because the line $\overline{p_{i-1}p_{i+1}}$ crosses the contour.]{
          \label{fig:concavePointsA}
          \includegraphics[width=.3\columnwidth]{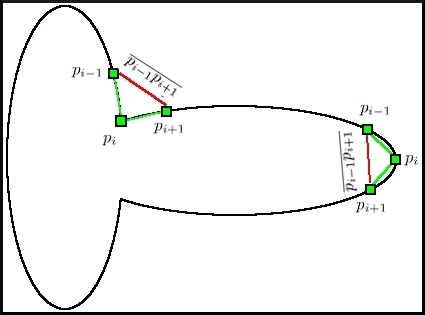}}
     \subfigure[Insertion of concave points. Found concave point (green) and inserted concave point (red).]{
          \label{fig:concavePointsB}
          \includegraphics[width=.3\columnwidth]{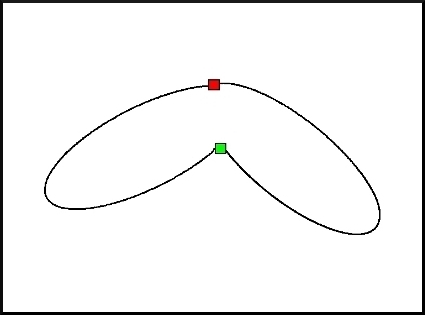}}
     \caption{Calculation of concave points and the insertion of points in a special case.}
\end{figure}

The approximated contour $PAC$ is divided at concave points to split touching cells.
These points are identified based on the angle of three consecutive points.
Given three points $p_{i-1}, p_{i}, p_{i+1}$, point $p_{i}$ is a concave point if the angle $\theta(i)$ (Equation \ref{eq:angle}) is between the minimum angle $\theta_{min}$ and the maximum angle $\theta_{max}$ (Equation \ref{eq:ang}).
In addition, to qualify a point as a concave point, the line $\overline{p_{i-1}p_{i+1}}$ should not cross the contour, as illustrated in Figure \ref{fig:concavePointsA}.
This second rule is needed to discard false concave points.

\begin{equation}
 \label{eq:angle}
 \theta(i) = \left\{ \begin{array}{ll}
 d_{i-1,i,i+1} & \textrm{, if } d_{i-1,i,i+1} < \pi \\
 \pi - d_{i-1,i,i+1} & \textrm{, otherwise}\\
 \end{array} \right.
\end{equation}
where $d_{i-1,i,i+1} = |a(p_{i-1}, p_{i}) -  a(p_{i+1}, p_{i})|$ and $a(p_{i}, p_{j}) = tan^{-1} \frac{(y_i - y_j)}{(x_i-x_j)}$.

\begin{equation}
 \label{eq:ang}
 \begin{array}{l}
   \theta_{min} < \theta(i) < \theta_{max} \\
   \overline{p_{i-1}p_{i+1}} \textrm{ should not cross the contour C.}
 \end{array}
\end{equation}

In some cases, touching between two or more cells has only one concave point $p_{c}$ that can be identified by the rules above.
In these cases, a new concave point is inserted at the opposite side of the identified concave point.
Following the assumption that the cells in the image have similar size, the position of the new concave point is the middle of the contour, considering the position $c$ of the single concave point equal to 0, as illustrated in Figure \ref{fig:concavePointsB}.
Another special case is the insertion of concave points at incomplete cells whose contour reached the image boundaries.
In these cases, concave points are inserted at the beginning and the end of the contour. An example can be seen in Figure \ref{fig:segmentationContourB}.

The concave points divide the contour into segments.
These segments are represented by $c_i = \{ p_{i1}, p_{i2}, ..., p_{is} \}$, where $s$ is the number of points of the segment $c_{i}$, $p_{i1}$ and $p_{is} $ are concave points.
If there are $m$ concave points, the contour is divided into $m$ segments such that $C = c_{1} \cup c_{2} \cup ... \cup c_{m}$.
Figure \ref{fig:segmentationContour}(c) shows an example of concave points and segments.

\subsection{Ellipse Processing}
Most cells can be modeled by an ellipse or a circle.
Thus, the purpose of this step is to model each contour segment with an ellipse.
These ellipses are processed in several steps that combine or divide them according to rules derived from prior knowledge of the cells.
For each contour segment $c_{i}$, an ellipse $e_{i}$ is fitted by an ellipse fitting algorithm.
Following \cite{Bai2009}, direct least square method \cite{Fitzgibbon1999} was used because it is computationally efficient and provides robust results even with noise and occlusions.
After ellipse fitting for each segment, the steps below are performed.

\subsubsection{Ellipse Selection}
The ellipses must satisfy two conditions to be selected.
First, the mean algebraic distance \cite{Fitzgibbon1999}, which measures the quality of the ellipse given the points, must be smaller than a threshold $disTh$.
Second, the ratio of the minor axis to major axis of the ellipse should be greater than a threshold $eTh$.
This second condition discards too slender ellipses.
The selected ellipses are used in the ellipse combination step, while the ellipses that were not selected are used in the last step (ellipse refinement).

\subsubsection{Ellipse Combination}
At this point, the cells are basically separated.
However, there may be segments belonging to the same cell erroneously separated by concave points misidentified.
As some cells do not have an ellipse shape or have a high mean algebraic distance error, the rules are also derived from the knowledge of the cells in the images \cite{Bai2009}.
These rules are described below in two cases.

Case 1: The simple touching of two cells is easily identified by the rules of the case 1.
These rules do not combine two ellipses whose touching is explicit.
As we are not interested in combining the ellipses, the distance of the center of the new ellipse $ec_{new}$ and the center of the two previous ellipses $ec_{i}$ and $ec_{j}$ should be greater than a threshold $dMinTh$, according to Equation \ref{eq:eqcase1}.

\begin{equation}
 \label{eq:eqcase1}
 \begin{array}{l}
   dist(ec_{i}, ec_{new}) > dMinTh \\
   dist(ec_{j}, ec_{new}) > dMinTh
 \end{array}
\end{equation}
where $dist(p,q)$ is the Euclidean distance of the points $p$ and $q$.

Threshold $dMinTh$ is easily determined using cell properties, usually close to the length of the minor axis of the smallest cell \cite{Bai2009}.
Another rule used in the case 1 says that two cells should be separated if the distance of the two previously cell centers is considerable, according to Equation \ref{eq:eqcase1_2}.

\begin{equation}
 \label{eq:eqcase1_2}
 dist(ec_{i}, ec_{j}) > [2.5, 4.0] dMinTh
\end{equation}

%
%

Case 2: Consider two segments $c_{i}$ and $c_{j}$ and their ellipses $e_{i}$ and $e_{j}$. 
Consider also, a segment $c_{ij} = c_{i} \cup c_{j}$ and its ellipse $e_{ij}$.
If the segments $c_i$ and $c_j$ belong to the same cell, the mean algebraic distance of the new ellipse $e_{ij}$ is probably smaller than the distances obtained by the two previous ellipses $e_i$ and $e_j$.
If this occurs, the segments should be combined.

The algorithm for combining ellipses is given in Algorithm 1.
\begin{algorithm}
  \label{algo_comb}
  \SetKwInOut{Input}{Input}
  \SetKwInOut{Output}{Output}
  \caption{Ellipse Combination Algorithm.}
  \Input{Segments $c_{i}$ and their ellipses $e_{i}$}
  \BlankLine

  \For{$i=1$ \KwTo $M$}{
   \For{$j=i+1$ \KwTo $M$}{
    $c_{ij} = c_{i} \cup c_{j}$\;
    Fit an ellipse $e_{ij}$ for $c_{ij}$\;

    \If{not Case\_1 and Case\_2} {
        Replace $e_{i}$ by $e_{ij}$\;
        Replace $c_{i}$ by $c_{ij}$\;
        Delete $e_{j}$ and $c_{j}$\;
        $G = G - 1$ and $i=1$\;
    }
   }
  }
\end{algorithm}

\subsubsection{Ellipse Refinement}
At this step, segments that have not been processed are used to refine the ellipses (e.g. segments with a small number of points and segments whose ellipse were not selected in the selection ellipse step).
For this, each unprocessed segment is concatenated with all existing segments and an ellipse is fitted for each combined segment.
After, the unprocessed segment belongs to the ellipse that provides the smaller mean algebraic distance and is still acceptable under the terms of the ellipse selection step.

\Section{Experiments and Results}
\label{sec:resultados}
We have conducted two sets of experiments to evaluate the proposed method.
The first experiment aims to validate the proposed method using images annotated by specialists. In the second series of experiments, the proposed method was applied to different types of cells, ranging from bacteria to blood cells.

First, experiments were performed on annotated images of biofilms of Lactobacillus paracasei, bacteria in the human mouth.
The motivation for using these images is the necessity to quantify the area and the number of bacteria before and after the use of chemical solutions.
The chemical solutions aim to reduce the number of bacteria, as there is an unrestricted formation of biofilms on the tooth surface, which is associated with the occurrence of diseases like dental caries.

For both experiments, image segmentation was performed by k-means algorithm with $k=3$ because the images contains, besides the background, two types of bacteria.
The remaining parameters were empirically adjusted as follows.
In the contour processing step, the threshold $dTh$ was set at $3.5$.
Due to the small size of cell in relation to the image size, the threshold $dTh$ was set to a low value that corresponds to the maximum polygon approximation error in pixels.
To calculate the concave points, the minimum angle $\theta_{min}$ and the maximum angle $\theta_{min}$ were set on $35^{\circ}$ and $155^{\circ}$, respectively.

The fitted ellipse for each segment must satisfy two constraints: mean algebraic distance should be less than $disTh$ and ratio between minor axis and major axis should be greater than $eTh$.
The two parameters were: $disTh = 0.03$, to allow certain robustness in the ellipse fitting, and $eTh = 0.2$, to restrict very elongated ellipses.
Finally, in the ellipse combination step, the threshold $dMinTh$ was $4$, which corresponds to the minor axis of the smallest cell in the images of training.
Each object has different properties, so the parameters used in the proposed method should be adjusted according to a priori knowledge of the object.

\begin{figure}[!htbp]
     \centering
     \subfigure[Proposed method results.]{
          \includegraphics[width=0.24\columnwidth]{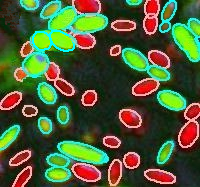}}
     \subfigure[Image marked by a specialist.]{
          \includegraphics[width=0.24\columnwidth]{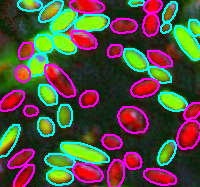}}
     \subfigure[Proposed method results.]{
          \includegraphics[width=0.24\columnwidth]{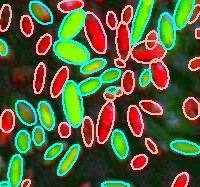}}
     \subfigure[Image marked by a specialist.]{
          \includegraphics[width=0.24\columnwidth]{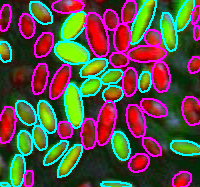}}
     \caption{Results for images with high density of touching cells.}
     \label{fig:resultados}
\end{figure}

In the first experiment, the proposed method was performed on 167 images with high density of bacteria.
Below, experimental results in this application are presented and discussed.
The proper detection of touching objects is one of the main difficulties of the methods of the literature.
However, this task is necessary for images with high density of cells.
The correct identification of cells provides estimates closer to reality and thus, more reliable results are obtained.
In Figure \ref{fig:resultados}, results for images with touching bacteria are presented.
The figures \ref{fig:resultados}(a) and \ref{fig:resultados}(c) correspond to the results obtained by the proposed method, while the other figures were marked by a specialist to validate the method.
Despite the large number of touching cells, proposed method achieves similar results to specialist in both images.

For the same images, the count of cells was performed by the proposed method and faced with the count carried out by three specialists (Table \ref{tab:results}).
We note that, even between specialists, there are differences due to the bias of each specialist.
Nevertheless, results obtained by the proposed method were similar to the average among specialists in both images.

\begin{table}[!htbp]
	\caption{Counting of bacteria carried out by the proposed method and three specialists.}
	\begin{center}
	\begin{tabular}{c|c|c|c|c}
	\hline
	& \multicolumn{2}{c|}{Figure \ref{fig:resultados}(a)} & \multicolumn{2}{c}{Figure \ref{fig:resultados}(c)}\\
	\hline 
	Detection Method & Bacteria 1 & Bacteria 2 & Bacteria 1 & Bacteria 2 \\
	\hline	\hline
	Specialist 1 & 23 & 24 & 24 & 29 \\
	Specialist 2 & 22 & 26 & 23 & 31 \\
	Specialist 3 & 21 & 25 & 22 & 30 \\
	Specialist Average & 22 & 25 & 23 & 30 \\
	Proposed Method & 23 & 25 & 24 & 31 \\
	\hline
	\end{tabular}		
	\label{tab:results}
	\end{center}
\end{table}

\begin{figure}[!htbp]
     \centering
     \subfigure[$e = 17.13$.]{
          \includegraphics[width=0.48\columnwidth]{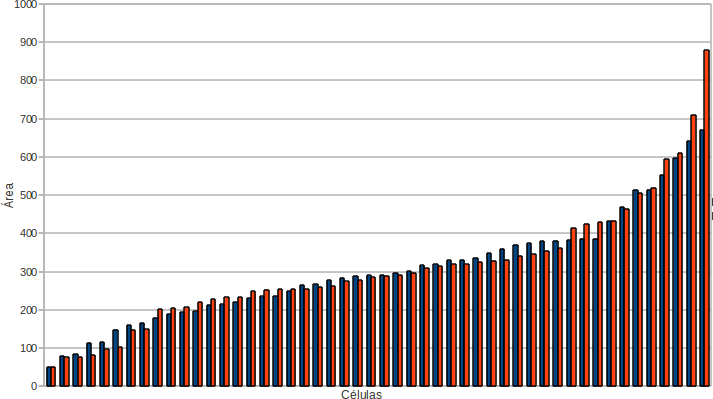}}
     \subfigure[$e = 29.52$.]{
          \includegraphics[width=0.48\columnwidth]{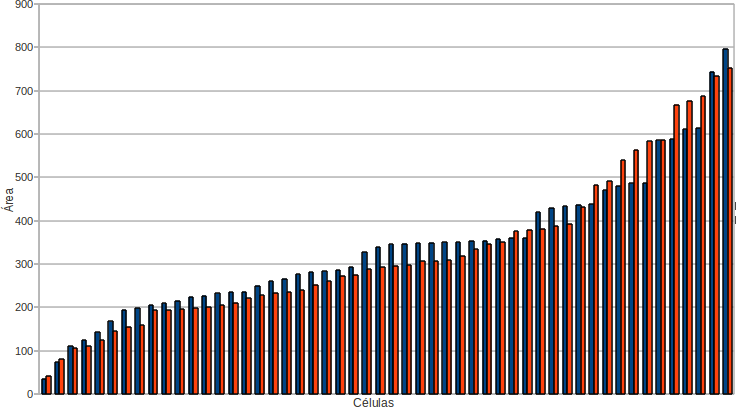}}
     \caption{Sorted bacteria areas for two images calculated by the proposed method and by a specialist.}
     \label{fig:resultados_hist}
\end{figure}

Figure \ref{fig:resultados_hist} presents a comparison of bacteria areas calculated by the proposed method and manual tracing in the two images.
For both images, bacteria areas were sorted to create the plot.
As can be seen, the method also obtains good results with respect to the area, which can be corroborated by the average error in pixels of $17.13$ and $29.52$ for images \ref{fig:resultados_hist}(a) and \ref{fig:resultados_hist}(b), respectively.

\begin{figure}[!htbp]
     \centering
     \includegraphics[width=0.8\columnwidth]{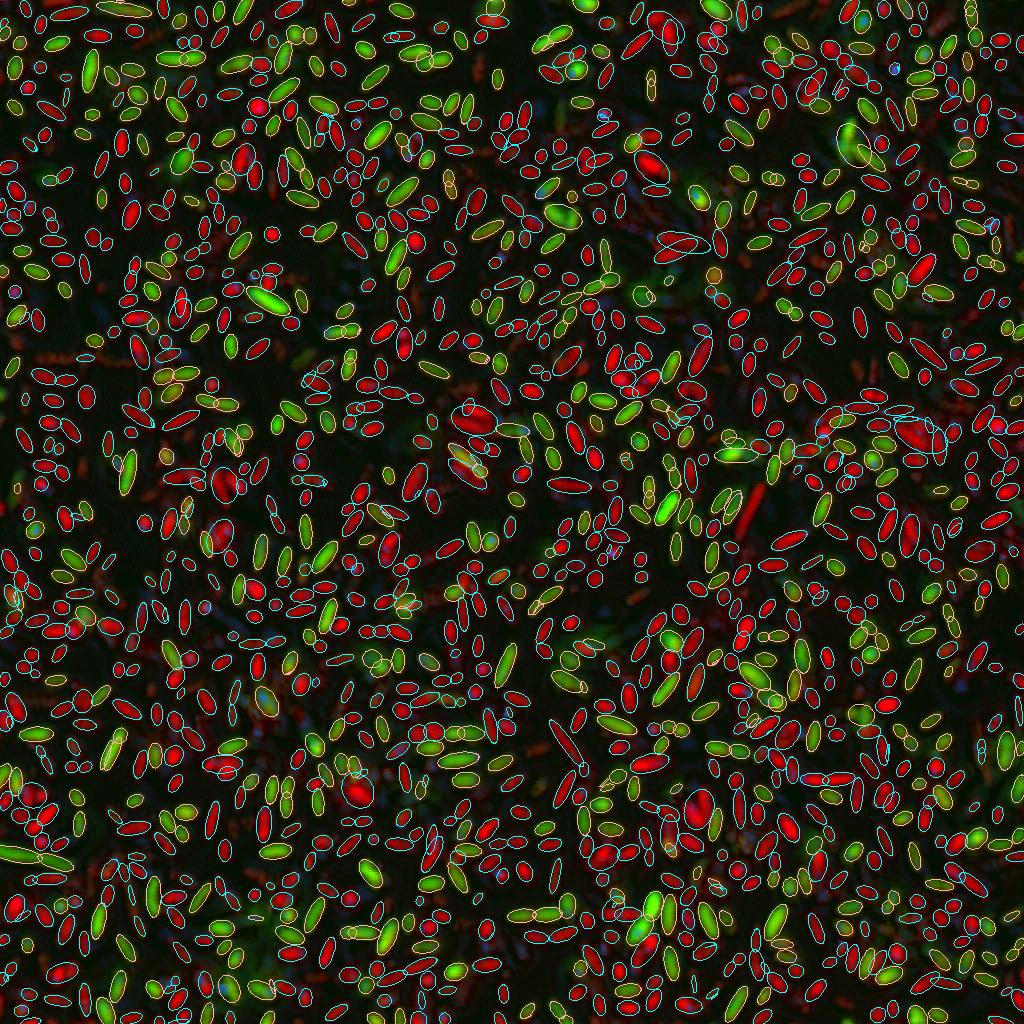}
     \caption{Results for an image with high concentration of two types of bacterias.}
     \label{fig:resultados2}
\end{figure}

\begin{figure}[!htbp]
     \centering
     \subfigure[Histogram of area for the green bacterias.]{
          \includegraphics[width=0.45\columnwidth]{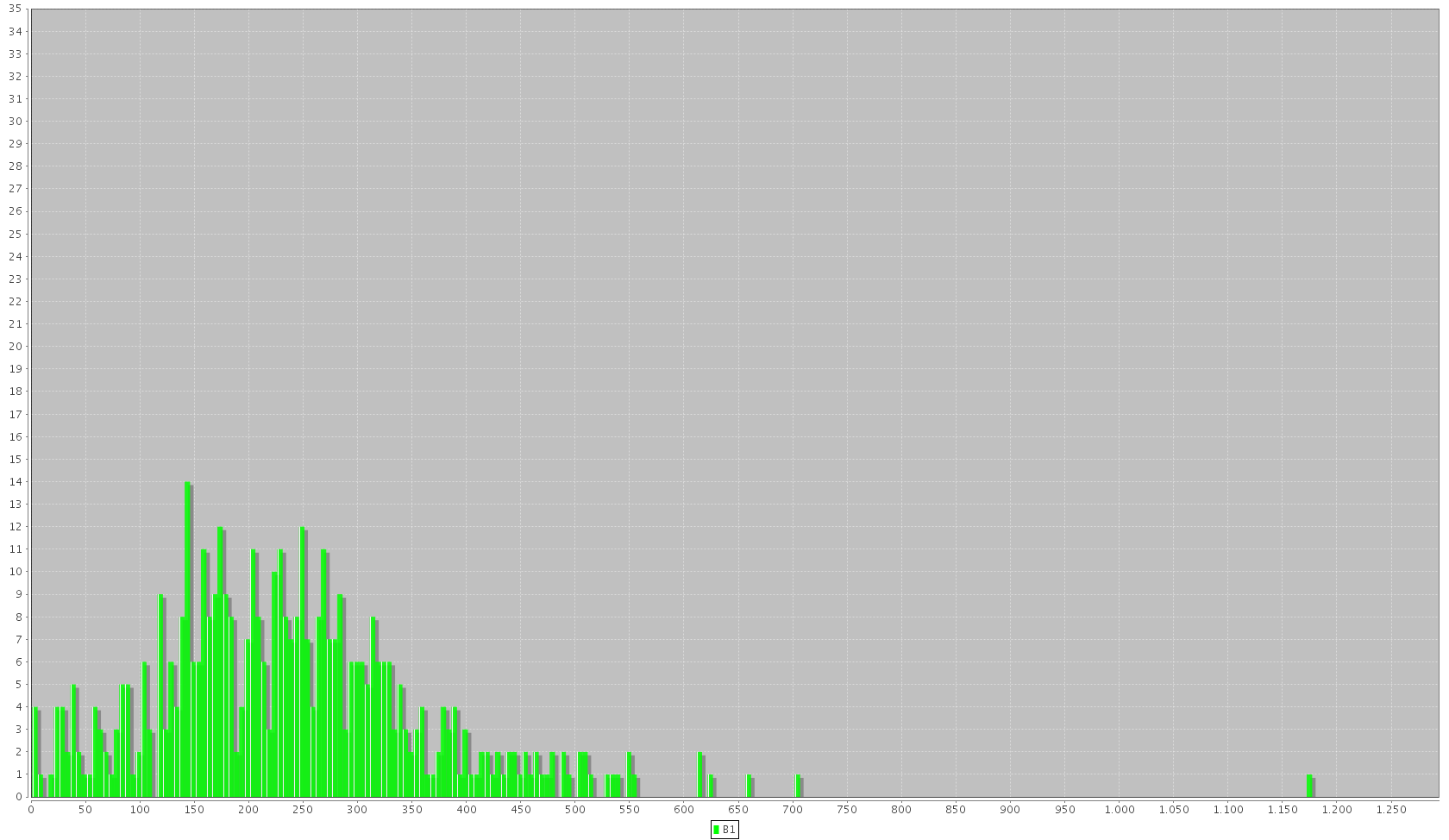}}
     \subfigure[Histogram of area for the red bacterias.]{
          \includegraphics[width=0.45\columnwidth]{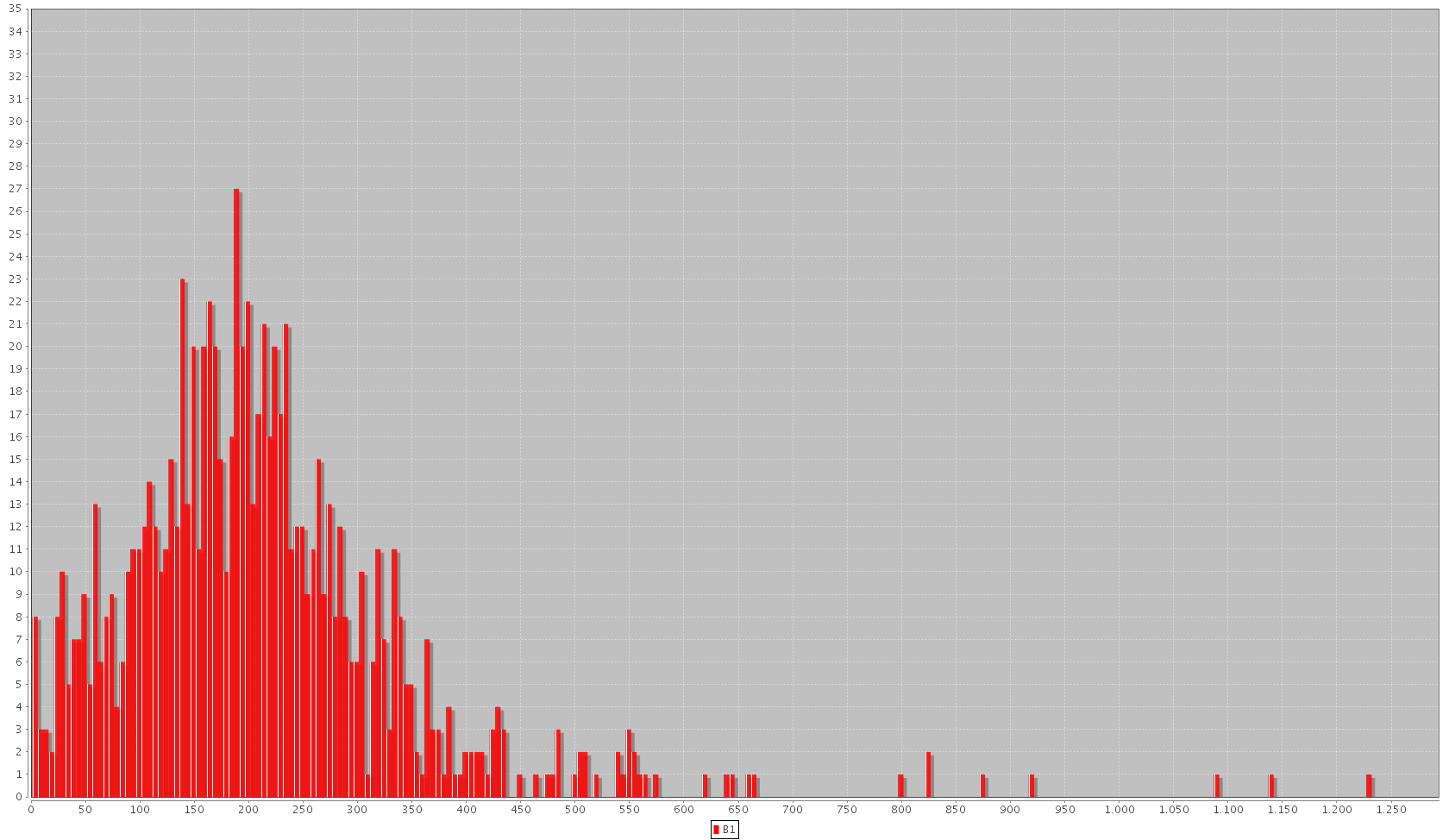}}
     \caption{Histogram of area for both types of bacterias.}
     \label{fig:hist_bacterias}
\end{figure}

In the second set of experiments, the proposed method was applied to three species of cells.
Figure \ref{fig:resultados2} shows results for a complete image of bacteria cells used in the earlier experiments. Despite the large amount of bacteria, the results are interesting because the process is fully automated. In Figure \ref{fig:hist_bacterias}, histogram of area for each type of bacteria is presented. These histograms can be used for evaluating chemical solution that combats mouth diseases.

\begin{figure}[!htbp]
     \centering
     \includegraphics[width=0.8\columnwidth]{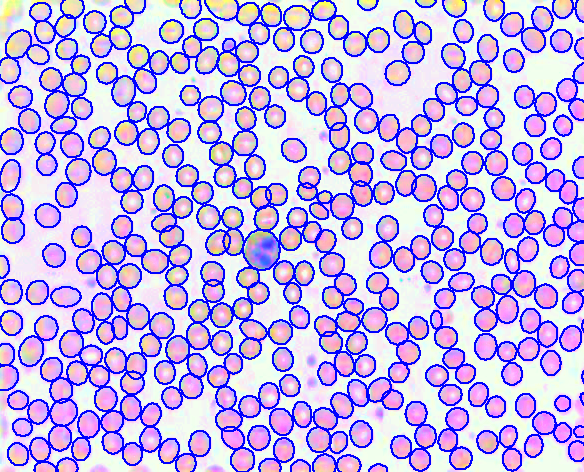}
     \caption{Results of detection for blood cell image.}
     \label{fig:bloodcell}
\end{figure}

\begin{figure}[!htbp]
     \centering
     \includegraphics[width=0.7\columnwidth]{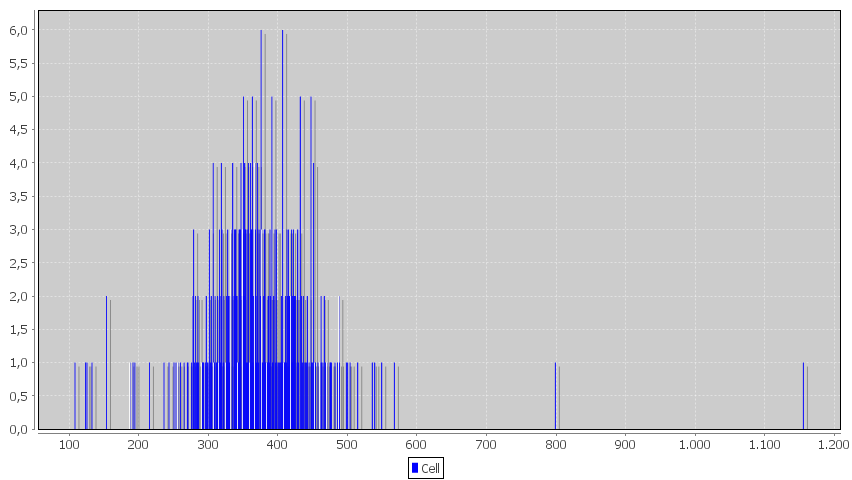}
     \caption{Histogram of area for the blood cells.}
     \label{fig:hist_blood}
\end{figure}

To evaluate the proposed method in other images, experimental results for blood cells are presented in Figure \ref{fig:bloodcell}. This image contains only one type of cell, which histogram of area is presented in Figure \ref{fig:hist_blood}. We have found that this method is a very useful technique for various type of cells, since it has the advantage of predict the shape of cells occluded due to the touching, as can be seen in Figure \ref{fig:bloodcell}.

\begin{figure}[!htbp]
     \centering
     \includegraphics[width=0.7\columnwidth]{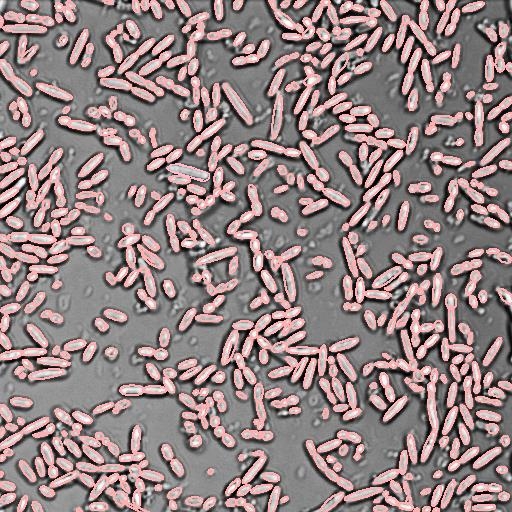}
     \caption{Results for mouth bacteria in another stage.}
     \label{fig:bact}
\end{figure}

\begin{figure}[!htbp]
     \centering
     \includegraphics[width=0.7\columnwidth]{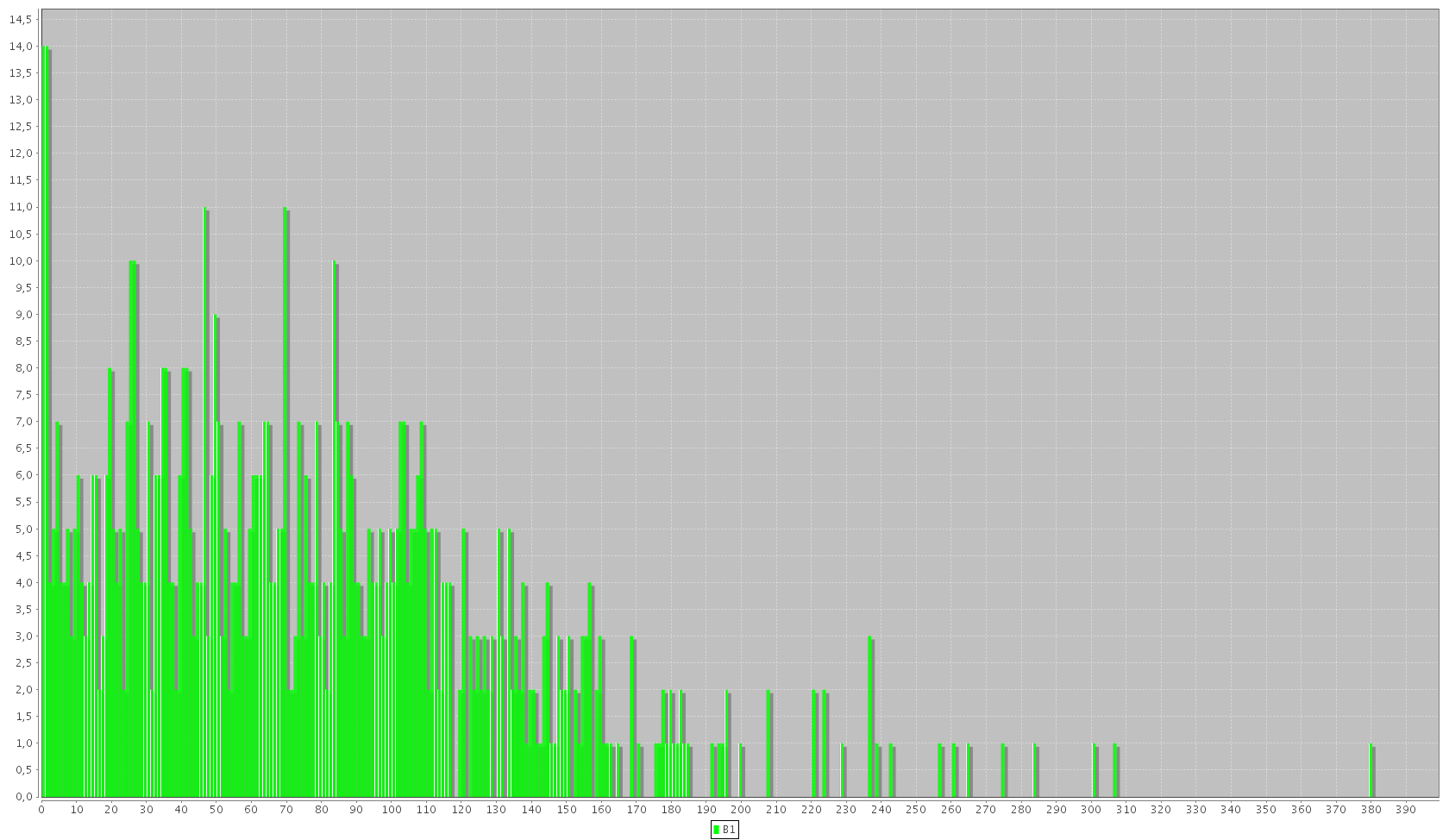}
     \caption{Histogram of area for mouth bacteria.}
     \label{fig:histbact}
\end{figure}

Finally, the proposed method was applied to mouth bacteria in another stage. The results of detection are presented in Figure \ref{fig:bact}. Although the bacteria in this stage have more elongated shape, the proposed method achieves good results of detection with respect to area and number of bacteria in the image. The histogram of area is presented in Figure \ref{fig:histbact}.

Besides the excellent results in detection of cells, the proposed method is also efficient in processing time. On average for 10 images with $1024 \times 1024$ pixels and high density of bacteria, the method took $553$ milliseconds on a computer Intel Quad Core 2.33GHz CPU and 3 GB RAM.

\Section{Conclusion}
\label{sec:conclusao}
This paper proposed a new approach for identifying and quantifying cells in images.
The proposed method consists of image segmentation based on k-means algorithm and an important step of contour processing to separate touching cells. Promising results have been obtained on three types of cells.
Experimental results indicate that the proposed method achieves detection performance comparable to detection performed by specialists.
In addition, our method makes the detection of cells feasible and simple, which results in an efficient and low cost implementation.

The proposed method is able to successfully handle a wide range of types of cells.
As part of the future work, we plan to focus on investigating the performance of the method on artificial images.
Another research issue is to evaluate other strategies to segment the images based on the watershed algorithm.

\section*{Acknowledgments}
The authors would like to thank Dr. Luis E. Ch\'{a}vez de Paz who provided images of cells.
WNG was supported by CNPq grants 142150/2010-0.
OMB was supported by CNPq grants 306628/2007-4 and 484474/2007-3.

\renewcommand\refname{References}


\end{document}